# Enhancing Multirotor Drone Efficiency: Exploring Minimum Energy Consumption Rate of Forward Flight under Varying Payload

Ayush Patnaik* and Nicolas Michel[†]
*University of California, Davis, California - 95616*

Xinfan Lin[‡]
*University of California, Davis, California - 95616*

**Multirotor unmanned aerial vehicle is a prevailing type of aircraft with wide real-world applications. Energy efficiency is a critical aspect of its performance, determining the range and duration of the missions that can be performed. In this study, we show both analytically and numerically that the optimum of a key energy efficiency index in forward flight, namely energy per meter traveled per unit mass, is a constant under different vehicle mass (including payload). Note that this relationship is only true under the optimal forward velocity that minimizes the energy consumption (under different mass), but not under arbitrary velocity. The study is based on a previously developed model capturing the first-principle energy dynamics of the multirotor, and a key step is to prove that the pitch angle under optimal velocity is a constant. By employing both analytical derivation and validation studies, the research provides critical insights into the optimization of multirotor energy efficiency, and facilitate the development of flight control strategies to extend mission duration and range.**

## I. Introduction

Multirotor unmanned aerial vehicles (UAVs) have emerged as a versatile tool in various applications, ranging from aerial photography and inspections, to intelligence, surveillance, and reconnaissance (ISR) missions, to package delivery.

Despite their widespread use, the limited endurance of these drones, typically around 30 minutes, poses a significant challenge and restricts their capabilities [1]. With the global market for commercial drones projected to surge to $12.7 billion by 2025 [2], the demand for extended flight endurance and range becomes increasingly critical.

The modeling and characterization of the energy dynamics and consumption of multirotors is important to optimize their energy performance. This has been achieved through modeling efforts which are either based on physical flight dynamics, or field measurements data. Noteworthy studies in this regard include those by [1], [3], [4], [5], [6] and [7]. Based on the modeling studies, extensive research has been devoted to identifying optimal flight conditions/factors that minimize energy consumption. One of the most important flight dynamic variables is the horizontal cruising velocity. Various studies such as [1], [8], [9] have highlighted the impacts of velocity on energy consumption. For instance, in [9], experimental measurement data are used to quantify the difference in energy consumption caused by horizontal velocity, while several model-based works, including those by [10], [11], [12] among others, study the relationship using model simulation. These works demonstrate the existence of an optimal velocity due to the trade-off of various factors, such as inflow aerodynamic effects and body drags, and hence velocity control has become an important topic in energy-optimal flight planning and control [13, 14].

Meanwhile, the impact of mass on energy consumption and energy-optimal flight behavior has not been systematically investigated. This gap is important to address, especially in contexts such as drone delivery and air mobility, where mass varies frequently and significantly. Prior works in [12], [15], [16] have generated some numerical results through simulations and experiments, showing how the energy consumption/power versus velocity curve changes under different mass. The findings indicate that energy consumption increases with mass, and interestingly, the optimal velocity also increases, barring any changes in the drag coefficient. There have also been works aiming at developing formula for energy consumption computation, which quantifies the relationship between energy efficiency, velocity, and vehicle mass. A key metric for quantifying energy efficiency is the energy consumption rate per unit distance, i.e. Energy Per Meter (EPM). An initial study by [17] derived the power consumed using a lift-to-drag ratio (over-)simplified as a function of drone velocity, from which the energy consumed for steady flight over distance is calculated by dividing the power with the speed. Other models like [18] provide an energy consumption model only considering hovering conditions, with the assumption that the energy consumption in horizontal steady-state flight is approximately the same. However, experimental data obtained using a Hexa-B drone showed the calculated energy consumption of the forward flight to be excessively high for both payload and no payload cases. The main issue with these existing studies is that the computed EPM overlooks the complexities of the aerodynamic effects during forward flight, giving inaccurate results due to oversimplifications. A detailed comparison of these studies and our results will be provided in Section V.

---

*Graduate Student Researcher, Department of Mechanical and Aerospace Engineering, ayupat@ucdavis.edu and AIAA Student Member
[†]Graduate Student Researcher, Department of Mechanical and Aerospace Engineering, nicmichel@ucdavis.edu
[‡]Associate Professor, Department of Mechanical and Aerospace Engineering, lxflin@ucdavis.edu

The primary contribution of this work is an in-depth analytical and numerical study of the energy efficiency, velocity, and vehicle mass relationship. We have made several notable discoveries for the first time, including that (1) EPM/m remains constant with respect to mass under the energy-optimal velocity, indicating a direct proportionality between optimal EPM and mass; (2) the optimal velocity is proportional to the square root of the mass; and (3) the optimal pitch angle remains unchanged with respect to mass. These results will be presented based on both numerical simulations and theoretical analyses, taking into account the multiphysical dynamics of multirotor flight, including propeller aerodynamics, inflow momentum, and airframe rigid body kinetics. The findings not only enhance the theoretical understanding of energy-optimal flight dynamics, but also offer practical insights applicable to flight control for practitioners. A main advancement of the state of art is that we demonstrate that the linear relationship between mass and EPM only holds under energy-optimal velocity, but not under any other general velocity. The latter is a commonly seen assumption/conclusion in existing literature, which is inaccurate due to oversimplified energy consumption models derived from hovering or quasi-hovering conditions.

## II. Modeling of Multirotor Energy Dynamics

A system-level model of a multirotor is used in this study, consisting of the aerodynamics of the propeller-rotor assembly, the electro-mechanical dynamics of the motor and electronic speed controller (ESC), the electrical dynamics of the battery [19], and the rigid body dynamics of the vehicle, as shown in Fig. 1. The low-level control signals are the Pulse-Width Modulation (PWM) commands sent to the motors by the flight controller. These commands are translated by ESC to motor input voltage as a fraction of the battery voltage sourced to ESC. In response to the PWM commands, the motor draws current and rotates at a certain angular velocity, which is also influenced by the torque load of the propeller. The motor drives the rotation of the propeller, which generates torque and thrust according to the blade element theory. The thrust and torque are then used in the rigid body dynamics model to calculate the motion of the multirotor, including the planar and perpendicular inflow velocities of each propeller, which in return affect the thrust and torque generation through propeller aerodynamics. The torque is also fed back to the motor, determining the motor speed and current, which are then looped back to the ESC to determine the current drawn from the battery. The model calculates the power consumption by multiplying the battery current with the voltage. In the following subsections, we will briefly discuss propeller aerodynamics and rigid body kinetics, which are the subsystem dynamics critical to the derivation and analysis in this work. Detailed information regarding the derivation and parameterization of the subsystem and overall models can be found in [10].

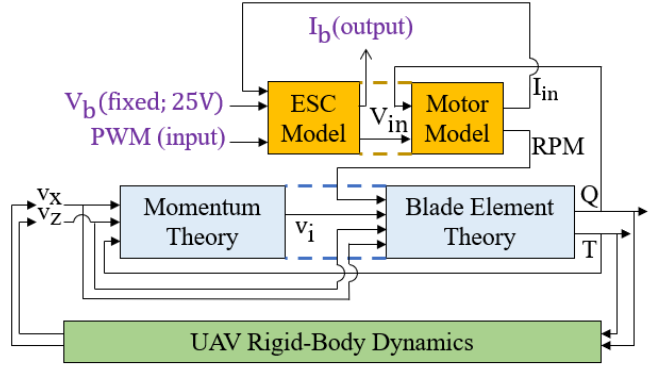

Fig. 1  Block Diagram of Model Architecture

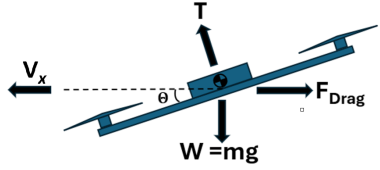

Fig. 2  Multirotor Free Body Diagram

## A. Key Subsystem Dynamics

*1. Rigid Body Kinetics*

The motion of the vehicle is modeled based on the rigid body dynamics of the airframe. In this work, 2D dynamics are considered, which are sufficient for horizontal flight,

$$\ddot{X} = \sum_{j=1}^{N_p} \frac{T_j \sin(\Theta)}{m} - \frac{C_{BD} V_x |V_x|}{m}$$
$$\ddot{Z} = \sum_{j=1}^{N_p} \frac{T_j \cos(\Theta)}{m} - g \qquad (1)$$
$$\ddot{\Theta} = \frac{\tau_\Theta}{J_\Theta} = \sum_{j=1}^{N_p} \frac{L_{\Theta,j} T_j}{J_\Theta}$$

where $X$ and $Z$ are the horizontal and vertical positions, $V_x$ and $V_z$ are the horizontal and vertical velocities of the vehicle center of mass in the global frame and $\Theta$ is the pitch angle. In addition, $\Sigma T_j$ is the sum of thrusts of all rotors computed by the propeller model (with each rotor indexed by subscript $j$ and total number of rotors $N_p$), $C_{BD} V_x |V_x|$ is the body drag force with $C_{BD}$ as the body drag coefficient, $\tau_\Theta$ is the total pitch-axis torque on the vehicle generated by the thrusts, $J_\Theta$ is the moment of inertia about the pitch axis, and $L_{\Theta,j}$ is the arm length of each rotor thrust to the pitch axis. The vertical drag associated with the vertical velocity $V_z$ is negligible in the steady-state horizontal flight, and hence not considered.

Finally, the velocities of each rotor (center) in the vehicle frame can be calculated as

$$v_{x,j} = V_x \cos(\Theta) + V_z \sin(\Theta)$$
$$v_{z,j} = -V_x \sin(\Theta) + V_z \cos(\Theta) + \dot{\Theta} x_j \qquad (2)$$

where $x_j$ is the x-axis position of the propeller relative to the vehicle center of mass.

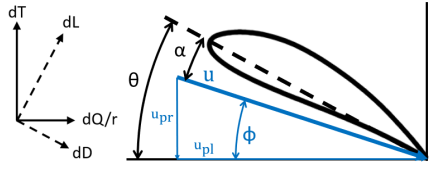

**Fig. 3   Geometries of a Blade Cross-section**

*2. Propeller Aerodynamics*

The propeller model uses the blade element and momentum theory to calculate the thrust $T$ and torque $Q$ generated by each propeller, with propeller angular velocity $\omega$, horizontal and vertical velocities, $v_x$ and $v_z$ computed by the rigid body kinetics as inputs. Noted that since all equations are related to a single propeller in this subsection, we dropped the subscript $_j$ of all notations for brevity.

An infinitesimal segment of the propeller blade are shown in Fig. 3, where the lift and drag force element $dL$ and $dD$ generate the thrust $dT$ and torque $dQ$ of the segment. The thrust and torque of the whole propeller blade can be computed by integrating $dT$ and $dQ$ along the blade length coordinate $r$ and averaged over one rotation cycle,

$$
\begin{aligned}
T &= \int_0^{2\pi} \int_{R_0}^{0.97R} dT\, d\psi\, dr/2\pi \\
&= \int_0^{2\pi} \int_{R_0}^{0.97R} (0.5 N_b \rho u_{pl}^2 c a (\theta - u_{pr}/u_{pl}))\, d\psi\, dr/2\pi \\
Q &= \int_0^{2\pi} \int_{R_0}^{R} dQ\, d\psi\, dr/2\pi \\
&= \int_0^{2\pi} \int_{R_0}^{R} (0.5 N_b r \rho u_{pl}^2 c (\phi a (\theta - u_{pr}/u_{pl}) + c_d))\, d\psi\, dr/2\pi,
\end{aligned} \quad (3)
$$

where $\psi$ represents the angular position of the blade along its rotating direction, $N_b$ is the number of blades in each propeller, $\rho$ is the air density, $c_d$ is the blade drag coefficient, $a$ is the blade lift coefficient factor, $c$ is the blade chord length, and $\theta$ is the twist angle of the blade. Integration is performed from the base of the blade $R_0$ to 97% of the tip $R$ instead of 100% to approximate tip loss. The planar inflow velocity $u_{pl}$, perpendicular inflow velocity $u_{pr}$, and inflow angle $\phi$ can be computed according to Fig. 3 as

$$
\begin{aligned}
u_{pl}(r,\psi) &= \omega r + v_x \sin(\psi), \quad u_{pr}(r) = v_i + v_z \\
\phi(r,\psi) &= \tan^{-1}(u_{pr}/u_{pl}).
\end{aligned} \quad (4)
$$

In addition, the rotor disk induced air velocity $v_i$, which is the downward velocity of the air imparted by propeller downwash, needs to be determined by applying the momentum theory to the propeller air stream. As depicted in Fig. 4, air enters the propeller airflow at point 1 with a relative velocity $v$ equal to the total UAV airspeed,

$$
v^2 = v_x^2 + v_z^2. \quad (5)
$$

The airflow undergoes acceleration perpendicular to the propeller disk plane, resulting in an induced velocity $v_o$ at the stream outlet (point 3) and $v_i$ at the disk (point 2), both assumed to be uniform across relevant cross-sections as shown in Fig. 4. By applying the principles of momentum and kinetic energy conservation, $v_o$ is determined to be twice of $v_i$. Subsequently, the relationship between the air mass flow rate $\dot{m}_{air}$ at the propeller, thrust $T$, and induced velocities can be established through the conservation of momentum [20],

$$
\begin{aligned}
T &= \dot{m}_{air} \cdot v_o = 2 \cdot \dot{m}_{air} \cdot v_i = 2\rho \pi R^2 v_i \sqrt{v_x^2 + (v_z + v_i)^2}, \\
T &= 2\rho \pi R^2 v_i \sqrt{(V_x \cos \Theta)^2 + (V_x \sin \Theta + v_i)^2}, \\
T &= 2\rho \pi R^2 v_i \sqrt{V_x^2 + v_i^2 + 2 V_x v_i \sin \Theta}.
\end{aligned} \quad (6)
$$

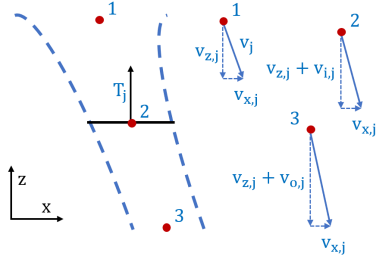

**Fig. 4  Schematic of Momentum Theory**

The above equations have also been simplified using the small-angle approximation and the conditions $u_{pr} \ll u_{pl}$ and $dD \ll dL$, which have been validated in [21]. Further $v_x$ and $v_z$ have been substituted with Eqn. 2 with steady-state horizontal flight conditions, $V_z = \dot{\Theta} = 0$. Rearranging (6) gives us

$$v_i^4 + 2V_x \sin\Theta v_i^3 + V_x^2 v_i^2 = \left(\frac{T}{2\rho\pi R^2}\right)^2, \qquad (7)$$

which is solved in a later section to determine $v_i$.

## B. Energy and Power Consumption

The power consumption of the UAV can be computed as the sum of the mechanical power of each propeller

$$P = \sum_{j}^{N_P} Q_j \omega_j, \qquad (8)$$

and the energy consumption is the integral of power over time,

$$E = \int_0^t P(\tau)d\tau. \qquad (9)$$

## III. Model-Based Energy Efficiency Simulations and Analysis

In this section, we present several important relationships regarding multirotor energy efficiency, velocity, and mass obtained from simulation studies using the previously introduced physics-based model. Studies are performed for steady-state horizontal flights under a range of constant velocities and different total mass (vehicle + payload). The horizontal drag coefficient $C_{BD}$ is treated as constant irrespective of payloads. The findings are outlined and discussed below.

Fig. 5 shows the energy efficiency metric, namely the energy consumption per meter traveled (EPM), under different vehicle mass and horizontal velocities for an Octorotor UAV under study. It is seen that in general, the energy consumed per meter is non-monotonic w.r.t. the forward velocity, $V_x$. These are classical results that can be explained by various tradeoff factors governing energy efficiency in steady forward flight, including velocity, aerodynamic factors, and drag resistance [21]. First, as the forward velocity increases, more distance is covered within unit time, contributing to the initial decrease in energy consumption per distance rate. Second, the aerodynamic effects induced by forward motion play a significant role. At low velocity $V_x$, the relative motion between the air and the propeller increases the inflow velocity component, i.e. $u_{pl}$ according to Eqn. (2), which will in turn increase the thrust production $T$ according to Eqn. (3) and thus improve energy efficiency. This effect is particularly pronounced at small pitch angles corresponding to low velocity. Third, as velocity increases, the quadratic growth of airframe body drag force begins to dominate, adversely affecting energy efficiency, especially at high velocities. Additionally, forward motion introduces another aerodynamic effect, wherein the velocity component perpendicular to the propeller rotation plane, i.e. $u_{pr}$, also increase as in Eqn. (2) and reduces thrust production according to Eqn. (3). This effect becomes more prominent under larger pitch angles, which further contributes to the increase of energy consumption at high velocity. As the mass of the vehicle increases (e.g. due to payload), it is not surprising to see that the energy consumption rate also increases as shown by the different curves in Fig. 5. However, an interesting yet nonintuitive observation is that the velocity associated with the optimal efficiency also increases with mass, indicating the need to fly heavier drone faster to save energy. Theoretical derivation will be provided in the subsequent section to prove this observation.

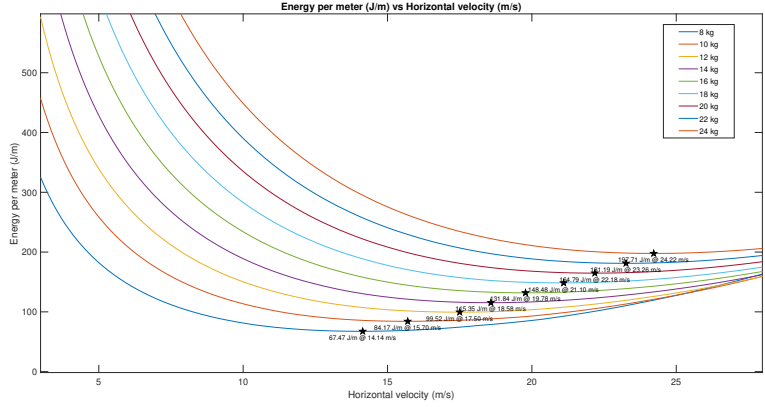

**Fig. 5** EPM vs Horizontal Velocity under Different Vehicle Mass

More interesting and related to the main topic of this paper, we further investigate the energy efficiency per unit mass. Noted that when considering different vehicle mass, a more relevant metric to evaluate energy efficiency is the EPM normalized by vehicle mass, i.e. energy consumption per unit distance traveled per unit mass. The EPM per mass curves are shown in Fig. 6 for the cases of different masses. It is seen that all curves attain the same minimum value, meaning that the optimal EPM per mass is actually the same under different vehicle mass, while the associated optimal velocity increases. Upon further investigation, we found that the pitch angle associated with the optimal velocity also remains constant under different vehicle masses as shown in Fig. 7. These intriguing findings, which have not been reported in the literature before, to the best of our knowledge, motivated our subsequent efforts to derive these constant relationships theoretically in the upcoming sections of our study.

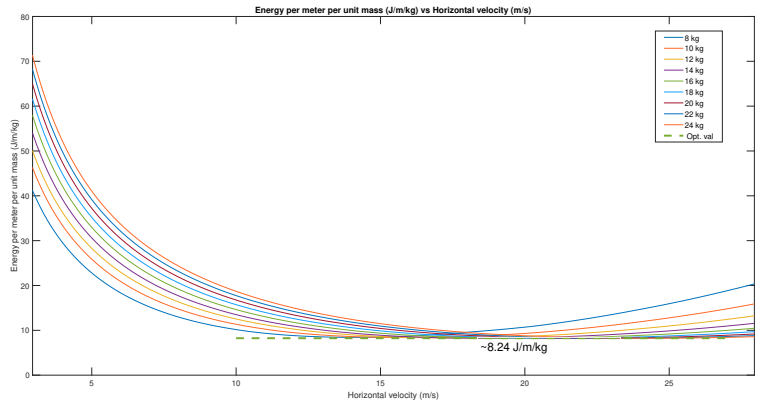

**Fig. 6** EPM per unit mass vs Horizontal Velocity under Different Vehicle Mass

## IV. Derivation of Theoretical Relationship

The goal of derivation is to mathematically prove that the optimal EPM per mass in constant-velocity steady-state forward flight is a constant under different vehicle mass, and the key step is to show that the pitch angle at the optimal EPM is independent of mass. The energy consumption per meter traveled (EPM) under constant horizontal velocity is

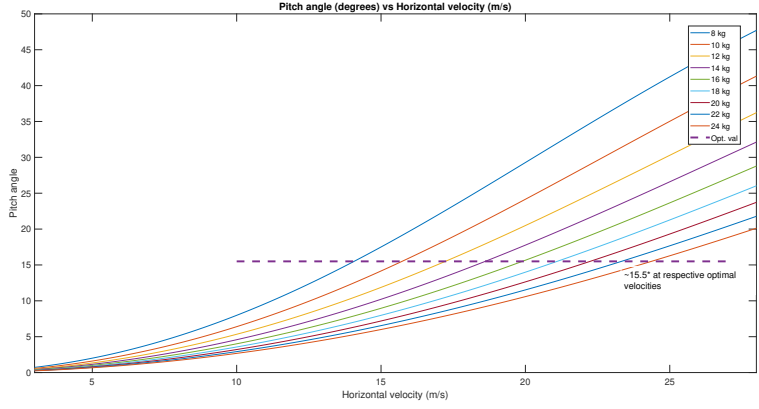

**Fig. 7 Pitch Angle vs Horizontal Velocity under Different Vehicle Mass**

formally defined as

$$EPM = \frac{Pt}{V_x t} = \frac{P}{V_x} = \frac{Q\omega}{V_x}, \tag{10}$$

due to the constant power consumption under steady-state flight both over time and among the propellers. In the following steps, we will show that under the energy-optimal velocity, the horizontal velocity ($V_x$), torque ($Q$), and angular velocity ($\omega$) can all be expressed as a certain power of mass ($m$) times a function of only pitch angle ($\Theta$) and aerodynamic constants. The results will then be used to prove the main results outlined above.

First, due to the force balance under constant-velocity horizontal flight shown in Fig. 2, we have

$$C_{BD} V_x^2 = T \sin \Theta, \quad T = \frac{mg}{\cos \Theta}, \tag{11}$$

which gives

$$V_x = \sqrt{\frac{mg}{C_{BD}} \tan \Theta} = \sqrt{m} V_x'(\Theta), \tag{12}$$

where

$$V_x'(\Theta) = \sqrt{\frac{g}{C_{BD}} \tan \Theta}, \tag{13}$$

is the desired mass-independent function of pitch angle, $\Theta$.

Then, based on the Blade Element Momentum Theory in Eqn. (3) and (4), thrust $T$ and torque $Q$ can be re-written as

$$T = \left[ B_{T1} \omega^2 + B_{T2} \frac{V_x^2}{2} - B_{T3} \omega (V_x \sin \Theta + v_i) \right] \tag{14}$$

with constants

$$\begin{aligned} B_{T1} &= \frac{N_p N_b \rho a}{2} \int_{R_0}^{R} r^2 c(r) \theta(r) dr \\ B_{T2} &= \frac{N_p N_b \rho a}{2} \int_{R_0}^{R} c(r) \theta(r) dr \\ B_{T3} &= \frac{N_p N_b \rho a}{2} \int_{R_0}^{R} r c(r) dr, \end{aligned} \tag{15}$$

and

$$Q = C_{Q1} \omega^2 + C_{Q2} V_x^2 + C_{Q3}(V_x \sin \Theta + v_i)\omega - 2 C_{Q2}(V_x \sin \Theta + v_i)^2 \tag{16}$$

with constants

$$\begin{aligned} C_{Q1} &= \frac{N_p N_b \rho c_d}{2} \int_{R_0}^{R} r^3 c(r) dr \\ C_{Q2} &= \frac{N_p N_b \rho c_d}{4} B_{T3} \\ C_{Q3} &= \frac{N_p N_b \rho a}{2} B_{T1}. \end{aligned} \tag{17}$$

Noted that the thrust and torque of all propellers are equal under the steady-state flight, and hence the total thrust and torque are those of each propeller times $N_p$.

In order to obtain the desired expression for $Q$, we need to solve for the propeller-induced velocity $v_i$ which is a quartic equation w.r.t $v_i$, using Eqns. 7 and 11,

$$v_i^4 + 2(V_x \sin \Theta) v_i^3 + V_x^2 v_i^2 = \left(\frac{mg}{2N_p \rho \pi R^2 \cos \Theta}\right)^2. \tag{18}$$

After a lengthy process of solving the quartic equation, the solution is obtained as

$$v_i = \sqrt{m} v_i'(\Theta) \tag{19}$$

where

$$v_i'(\Theta) = \sqrt{\frac{g}{C_{BD}} \tan \Theta \left( \frac{\sin \Theta}{2} + S_2' + \sqrt{-\frac{S_2'^2}{4} + \frac{3 \sin^2 \Theta - 2}{16} + \frac{\sin \Theta \cos^2 \Theta}{16 S_2'}} \right)} \tag{20}$$

Here, $v_i'$ is the desired function of pitch angle $\Theta$ and aerodynamic constant $S_2'$ without dependence on mass $m$. The complete derivation procedures for $v_i'$ and the form of $S_2$ will be provided in the appendix.

The next step is to get a similar expression for the propeller angular velocity $\omega$. By substituting the thrust $T$ from force balance in Eqn. (11) into Eqn. (14), we obtain a quadratic equation w.r.t $\omega$, as

$$B_{T1}\omega^2 - B_{T3}\sqrt{m}\left(\sqrt{\frac{g}{C_{BD}} \tan \Theta} \sin \Theta + v_i'\right)\omega + mg\left(\frac{B_{T2} \sin \Theta - 1}{2 C_{BD} \cos \Theta}\right) = 0 \tag{21}$$

which gives the solution:

$$\omega = \sqrt{m}\omega'(v_i', \Theta)$$
$$\omega'(v_i', \Theta) = \sqrt{\frac{g}{C_{BD}} \tan \Theta} \frac{B_{T2}}{2 B_{T1}} \left( v_i' + \sin \Theta + \sqrt{(v_i' + \sin \Theta)^2 + \frac{4 B_{T1}}{B_{T2}^2}\left(\frac{C_{BD}}{\sin \Theta} - B_{T3}\right)} \right) \tag{22}$$

with $\omega'$ being a function of the pitch angle $\Theta$ and $v_i'$. Since $v_i'$ is only a function of $\Theta$ as previously derived in Eqn. (19), we can write $\omega$ as

$$\omega = \sqrt{m}\omega'(\Theta). \tag{23}$$

Next, the torque $Q$ from Eqn. (16) can be re-written by substituting the expressions for $V_x$ in Eq. (12), $v_i$ in Eq. (19), and $\omega$ in Eq. (23) as

$$Q = C_{Q1} m \omega'^2(\Theta) + C_{Q2} m V_x'(\Theta)^2$$
$$+ C_{Q3} m (V_x'(\Theta) \sin(\Theta) + v_i'(\Theta)) \omega'(\Theta) \tag{24}$$
$$- 2 C_{Q2} m \left( V_x'(\Theta) \sin(\Theta) + v_i'(\Theta) \right)^2$$

Noted that since $V_x'$, $v_i'$, and $\omega'$ are all functions of only $\Theta$, we can write $Q$ as

$$Q = m Q'(\Theta). \tag{25}$$

Now, based on Eqns (23), (25) and (12), the EPM in Eqn. (10) can be written as

$$EPM = \frac{Q\omega}{V_x} = \frac{mQ'(\Theta)\sqrt{m}\omega'(\Theta)}{\sqrt{m}V_x'(\Theta)} = m\frac{Q'(\Theta)\omega'(\Theta)}{V_x'(\Theta)}. \tag{26}$$

Finally, in order to find the minimum $EPM$ in constant-velocity horizontal flight, we take the derivative of $EPM$ w.r.t $\Theta$ and find the optimum at 0,

$$\frac{\partial EPM}{\partial \Theta} = 0. \tag{27}$$

which yields

$$\left.\frac{\partial\left(\frac{Q'(\Theta)\omega'(\Theta)}{V'_x(\Theta)}\right)}{\partial\Theta}\right|_{\Theta^*} = 0, \quad (28)$$

where the superscript $^*$ denotes the optimum. Since $V'_x$, $Q'$, and $\omega'$ are all functions of only $\Theta$ and independent of mass $m$ as shown previously, the solution $\Theta^*$ of the above equation will also be independent of $m$ and remain constant over changing mass $m$. Therefore, after solving for $\Theta^*$, the minimal $EPM$ in Eqn. (26) can be written as proportional to the mass $m$,

$$EPM^* = m\left(\frac{Q'(\Theta^*)\omega'(\Theta^*)}{V'_x(\Theta^*)}\right) = mC, \quad (29)$$

or equivalently, the optimal EPM per mass is a constant,

$$\frac{EPM^*}{m} = C, \quad (30)$$

where the constant $C$ is defined as

$$C = \frac{Q'(\Theta^*)\omega'(\Theta^*)}{V'_x(\Theta^*)}. \quad (31)$$

The above equations prove the main conclusion of the paper. It is noted that here we take the derivative of $EPM$ to the pitch angle $\Theta$ to find the optimal operating condition, which is the same as taking the derivative to velocity $V_x$, because under steady-state flight $V_x$ and $\Theta$ are in a one-to-one mapping as seen in Eqn. (12).

## V. Implications of Results

The results from our analysis yield profound implications across multiple facets of multirotor dynamics, control, and operation planning. First, in the realm of flight control, the finding that the energy-optimal pitch angle $\Theta^*$ remains constant despite variation in payload enables simple yet effective energy-efficient flight control protocol. Specifically, the energy-optimal velocity can be obtained as

$$V_x^* = \sqrt{\frac{mg}{C_{BD}}\tan\Theta^*}, \quad (32)$$

where the mass-invariant $\Theta^*$ only needs to be solved once based on the model, and then plugged into Eqn. (32) to obtain the $V_x^*$ under any mass $m$. Alternatively, practitioners can manage to measure/identify $V_x^*$ under one specific mass $m$ using experimental techniques, e.g. as in [9, 13], and then scale with $\sqrt{m}$ to obtain $V_x^*$ under other mass conveniently. Consequently, the burden on repeated experimental measurements and/or online adaptation would be significantly reduced. The formulation of a precise formula that correlates energy-optimal velocity with the square root of mass enriches our understanding of dynamic control adjustments needed for maximizing energy efficiency, offering nuanced insights into the interplay between velocity management and payload characteristics.

Second, for multirotor energy consumption prediction and energy source design/sizing, our finding of a linear relationship between optimal energy consumption rate and total vehicle mass represents a highly useful advancement. Specifically, the minimum energy consumption $E^*$ over a certain distance $L$ under the energy-optimal velocity $V_x^*$ is

$$E^* = EPM^* L = CmL, \quad (33)$$

which can be used to determine the minimum requirement on energy capacity needed for the specified mission distance. Conversely, given a fixed capacity of the multirotor onboard energy storage (e.g. battery) $E$, the maximum range $L^*$ is

$$L^* = \frac{E}{Cm}, \quad (34)$$

which specifies the maximum range capability under the available energy capacity. The constant $C$ can be either computed theoretically based on Eqn. (31) or identified from experiment measurement. These simple formula enables quick yet accurate (theoretically proven) evaluation of operation range or energy requirement.

Third, our finding also facilitates energy-oriented optimization and planning of multirotor operations, especially those involving payload distribution. One prominent example is drone delivery, with the goal of delivering a series of payloads $\{m_i\}_{i=1}^{N_m}$ to various destination nodes by going through routes consisting of the line segments between nodes $\{L_j\}_{j=1}^{N_L}$. By using the formula derived in this work, the optimal routing problem can be formulated as choosing and pairing $L_i$ with $m_i$ to minimize the total energy consumption,

$$\min_{\{(m_i, L_i)\}_{i=1}^{N_m}} E = \sum_i C(m_v + m_i) L_i, \tag{35}$$

where $m_v$ is the vehicle weight. There are two major benefits brought by the above problem formulation. First, the solution will attain the "absolute" energy optimality, i.e. both optimal velocity control and optimal routing, as the energy consumption formula in Eqn. (33) applies to the flights operated under the energy-optimal velocity over each segment. This is a significant difference from the common practice in literature, which often assigns an arbitrary constant velocity for the convenience of solving the problem. Second, the problem is very easy to solve, as it is in the form of linear integer programming. The simple formulation is not based on any assumption/simplification, but rather on strict theoretical derivation thanks to the main finding of this paper, where physics-based energy dynamics are considered and encoded in the pre-solved coefficient $C$. More interestingly, if the goal is only to find the optimal routing (without necessarily knowing the total energy consumption), the constant coefficient $C$ can be eliminated from the problem and does not even need to be solved for. The problem hence becomes a mass-weighted Traveling Salesman Problem, which can be solved by practitioners even without any knowledge of multirotor energy dynamics, yet still yielding accurate results. Again, it needs to be emphasized that all these advantages hinge on the condition that the multirotor is operated under the energy-optimal velocity which varies with the mass, but not any arbitrary velocity.

In addition, we also want to highlight the fundamental difference between the findings of this paper and some widely adopted existing practice in literature. Many works in literature directly assume a proportional relationship between vehicle mass and energy consumption rate $EPM$ under any horizontal velocity, e.g. in the form of

$$EPM = \frac{mg}{r(V_x)\eta} \tag{36}$$

in [7] and,

$$EPM = P/V_x = \frac{(g \sum_{k=1}^{3} m_k)^{\frac{3}{2}}}{\eta V_x \sqrt{2n\rho\varsigma}} \tag{37}$$

in [18], where $r(V_x)$ is the lift-to-drag ratio as a function of the velocity, $\eta$ is the lumped efficiency of the system, and $\varsigma$ being the spinning area of the rotor. The formulae, however, essentially indicate that $EPM$ over $m$ (or its power) is a constant under the same (arbitrary) velocity $V_x$, which is not true according to Fig. 6. Instead, our derivation result in Eqn. (30) shows that $EPM/m$ is a constant under the same (arbitrary) pitch angle $\Theta$. Since the $V_x$ to $\Theta$ mapping changes under different masses, $EPM/m$ will obviously not remain constant with respect to velocity. The true invariant flight state that governs the energy efficiency (normalized by mass) is the pitch angle, which in our opinion is an insight that advances the understanding of the fundamentals of multirotor energy dynamics.

## VI. Conclusion

This paper explores the fundamental relationship between energy efficiency, velocity, and vehicle mass of the multirotor UAV under the steady-state forward flight, through both simulation studies and theoretical derivation. Several notable insights have been made for the first time. First, the energy-optimal pitch angle remains unchanged with respect to mass. Second, the energy-optimal velocity is proportional to the square root of mass. Third, the minimal energy consumption rate normalized by mass, i.e. $EPM/m$, remains constant with respect to mass, indicating a direct proportionality between optimal EPM and mass. These insights not only deepen our understanding of the fundamental multirotor energy dynamics, but also pave the way for developing energy-efficient operational strategies. By leveraging these findings, stakeholders can enhance the robustness, efficiency, and effectiveness of multirotor operations, thereby unlocking new opportunities for innovation and excellence in the field.

## Appendix

Solution of Eqn. (18):

$$v_i^4 + 2(V_x \sin \Theta)v_i^3 + v^2 v_i^2 = \left(\frac{mg}{2\rho\pi R^2 \cos \Theta}\right)^2$$

$$V_x = \sqrt{\frac{mg}{C_{BD}} \tan \Theta}$$

$$v_i^4 + 2\sqrt{\frac{mg}{C_{BD}} \tan \Theta} \sin \Theta v_i^3 + \frac{mg}{C_{BD}} \tan \Theta v_i^2 - \left(\frac{mg}{2\rho\pi R^2 \cos \Theta}\right)^2 = 0$$

$$a = 1 \tag{38}$$

$$b = 2\sqrt{\frac{mg}{C_{BD}} \tan \Theta} \sin \Theta \tag{39}$$

$$c = \frac{mg}{C_{BD}} \tan \Theta = K_1 \tan \Theta \tag{40}$$

$$d = 0 \tag{41}$$

$$e = -(mg)^2 \cos^{-2} \Theta \left(\frac{1}{2\rho\pi R^2}\right)^2 \tag{42}$$

$$K_1 = \frac{mg}{C_{BD}} \tag{43}$$

$$K_2 = \left(\frac{C_{BD}}{2N\rho\pi R^2}\right)^2 \tag{44}$$

$$K_1^2 K_2 = \left(\frac{mg}{2N\rho\pi R^2}\right)^2 \tag{45}$$

$$v = \sqrt{K_1 \tan \Theta}$$

$$p = c - \frac{3b^2}{8} = K_1 \sin \Theta \left(\cos \Theta - \frac{1}{2} \sin^2 \Theta \cos^{-1} \Theta\right)$$

$$q = \frac{b^3}{8} - \frac{bc}{2} = -K_1^{3/2} \sin^{2.5} \Theta \cos^{0.5} \Theta$$

$$\Delta_0 = c^2 + 12e = K_1^2 \tan^2 \Theta \left(1 - 12K_2 \sin^{-2} \Theta\right)$$

$$\Delta_1 = 2c^3 + 27b^2 e - 72ce = 2K_1^3 \tan^3 \Theta \left(1 + 18K_2(2\sin^{-2} \Theta - 3)\right)$$

$$S_1 = \sqrt[3]{\frac{1}{2}(\Delta_1 + \sqrt{\Delta_1^2 - 4\Delta_0^3})} = K_1 \tan \Theta \sqrt[3]{1 + 36K_2 \sin^{-2} \Theta - 54K_2 + \frac{\sqrt{\Delta_1^2 - 4\Delta_0^3}}{2K_1^3 \tan^3 \Theta}}$$

$$\Delta_2 = \frac{\sqrt{\Delta_1^2 - 4\Delta_0^3}}{2K_1^3 \tan^3 \Theta} = \sqrt{108K_2(-1 + 27K_2 + (1 - 36K_2)\sin^{-2} \Theta + 8K_2 \sin^{-4} \Theta + 16K_2 \sin^{-6} \Theta)}$$

$$S_1 = K_1 \tan \Theta \sqrt[3]{1 + 36K_2 \sin^{-2} \Theta - 54K_2 + \Delta_2}$$

$$S_1 = \sqrt{K_1 \tan \Theta} \cdot S_1^*$$

$$S_1 = V_x \cdot S_1'$$

$$S_2 = \sqrt{\frac{-p}{6} + \frac{S_1}{12} + \frac{\Delta_0}{12 S_1}} = \sqrt{K_1 \tan \Theta} \sqrt{-\frac{1}{6} + \frac{\sin^2 \Theta}{4} + \frac{S_1}{12 K_1 \tan \Theta} + \frac{K_1 \tan \Theta}{S_1}\left(\frac{1}{12} - K_2 \sin^2 \Theta\right)}$$

$$S_2 = V_x \sqrt{-\frac{1}{6} + \frac{\sin^2 \Theta}{4} + \frac{S_1}{12 K_1 \tan \Theta} + \frac{K_1 \tan \Theta}{S_1}\left(\frac{1}{12} - K_2 \sin^2 \Theta\right)}$$

$$S_2 = V_x S_2^*$$

$$v_i = -\frac{b}{4} + S_2 + \frac{1}{2}\sqrt{-S_2^2 - \frac{p}{2} - \frac{q}{4 S_2}} = \sqrt{K_1 \tan \Theta}\left(\frac{\sin \Theta}{2} + \frac{S_2}{\sqrt{K_1 \tan \Theta}} + \frac{1}{2}\sqrt{\frac{-S_2^2}{K_1 \tan \Theta} + \frac{3 \sin^2 \Theta}{4} - \frac{1}{2} + \frac{\sqrt{K_1 \tan \Theta} \cos^2 \Theta \sin \Theta}{4 S_2}}\right)$$

$$v_i = \sqrt{K_1 \tan \Theta}\left(0.5 \tan \Theta + \frac{S_2}{\sqrt{K_1 \tan \Theta}} + \frac{1}{2}\sqrt{\frac{-S_2^2}{K_1 \tan \Theta} + \frac{3}{4}\sin^2 \Theta - \frac{1}{2} + \frac{1}{4}\sin \Theta \cos^2 \Theta \frac{\sqrt{K_1 \tan \Theta}}{S_2}}\right)$$

$$v_i = \sqrt{K_1 \tan \Theta} \cdot v_i^*$$

$$v_i = V_x v_i^*$$

$$v_i^*(\Theta) = \frac{\sin \Theta}{2} + S_2' + \sqrt{\frac{-S_2'^2}{4} + \frac{3 \sin^2 \Theta - 2}{16} + \frac{\sin \Theta \cos^2 \Theta}{16 S_2'}}$$

$$v_i'(\Theta) = \sqrt{\frac{g}{C_{BD}} \tan \Theta}\left(\frac{\sin \Theta}{2} + S_2' + \sqrt{\frac{-S_2'^2}{4} + \frac{3 \sin^2 \Theta - 2}{16} + \frac{\sin \Theta \cos^2 \Theta}{16 S_2'}}\right)$$

## References

[1] Bauersfeld, L., and Scaramuzza, D., "Range, Endurance, and Optimal Speed Estimates for Multicopters," *IEEE Robotics and Automation Letters*, Vol. 7, No. 2, 2022, pp. 2953–2960. https://doi.org/10.1109/LRA.2022.3145063.

[2] Laghari, A. A., Jumani, A. K., Laghari, R. A., and Nawaz, H., "Unmanned aerial vehicles: A review," *Cognitive Robotics*, Vol. 3, 2023, pp. 8–22. https://doi.org/https://doi.org/10.1016/j.cogr.2022.12.004, URL https://www.sciencedirect.com/science/article/pii/S2667241322000258.

[3] Kirschstein, T., "Comparison of energy demands of drone-based and ground-based parcel delivery services," *Transportation Research Part D: Transport and Environment*, Vol. 78, 2020, p. 102209. https://doi.org/https://doi.org/10.1016/j.trd.2019.102209, URL https://www.sciencedirect.com/science/article/pii/S1361920919309575.

[4] Murray, C. C., and Raj, R., "The multiple flying sidekicks traveling salesman problem: Parcel delivery with multiple drones," *Transportation Research Part C: Emerging Technologies*, Vol. 110, 2020, pp. 368–398. https://doi.org/https://doi.org/10.1016/j.trc.2019.11.003, URL https://www.sciencedirect.com/science/article/pii/S0968090X19302505.

[5] Poikonen, S., and Golden, B., "Multi-visit drone routing problem," *Computers & Operations Research*, Vol. 113, 2020, p. 104802. https://doi.org/https://doi.org/10.1016/j.cor.2019.104802, URL https://www.sciencedirect.com/science/article/pii/S0305054819302448.

[6] Stolaroff, J. K., Samaras, C., O'Neill, E. R., Lubers, A., Mitchell, A. S., and Ceperley, D., "Energy use and life cycle greenhouse gas emissions of drones for commercial package delivery," *Nature Communications*, Vol. 9, No. 1, 2018. https://doi.org/10.1038/s41467-017-02411-5, cited by: 314; All Open Access, Gold Open Access.

[7] Figliozzi, M. A., "Lifecycle modeling and assessment of unmanned aerial vehicles (Drones) CO2e emissions," *Transportation Research Part D: Transport and Environment*, Vol. 57, 2017, pp. 251–261. https://doi.org/https://doi.org/10.1016/j.trd.2017.09.011, URL https://www.sciencedirect.com/science/article/pii/S1361920917304844.

[8] Zeng, Y., Xu, J., and Zhang, R., "Energy Minimization for Wireless Communication With Rotary-Wing UAV," *IEEE Transactions on Wireless Communications*, Vol. 18, No. 4, 2019, pp. 2329–2345. https://doi.org/10.1109/TWC.2019.2902559.

[9] Di Franco, C., and Buttazzo, G., "Energy-aware coverage path planning of UAVs," *2015 IEEE International Conference on Autonomous Robot Systems and Competitions*, IEEE, 2015, pp. 111–117.

[10] Michel, N., Wei, P., Kong, Z., Sinha, A. K., and Lin, X., "Modeling and Validation of Electric Multirotor Unmanned Aerial Vehicle System Energy Dynamics," *eTransportation*, Vol. 12, 2022, p. 100173. https://doi.org/https://doi.org/10.1016/j.etran.2022.100173.

[11] Hong, D., Lee, S., Cho, Y. H., Baek, D., Kim, J., and Chang, N., "Least-Energy Path Planning With Building Accurate Power Consumption Model of Rotary Unmanned Aerial Vehicle," *IEEE Transactions on Vehicular Technology*, Vol. 69, No. 12, 2020, pp. 14803–14817. https://doi.org/10.1109/TVT.2020.3040537.

[12] Dukkanci, O., Kara, B. Y., and Bektaş, T., "Minimizing energy and cost in range-limited drone deliveries with speed optimization," *Transportation Research Part C: Emerging Technologies*, Vol. 125, 2021, p. 102985. https://doi.org/https://doi.org/10.1016/j.trc.2021.102985, URL https://www.sciencedirect.com/science/article/pii/S0968090X21000206.

[13] Wu, X., and Mueller, M. W., "In-flight range optimization of multicopters using multivariable extremum seeking with adaptive step size," *2020 IEEE/RSJ International Conference on Intelligent Robots and Systems (IROS)*, IEEE, 2020, pp. 1545–1550.

[14] Michel, N., Wei, P., Kong, Z., and Lin, X., "Energy-Optimal Unmanned Aerial Vehicles Motion Planning and Control Based on Integrated System Physical Dynamics," *Journal of Dynamic Systems, Measurement, and Control*, Vol. 145, No. 4, 2023, p. 041002. https://doi.org/10.1115/1.4056534, URL https://doi.org/10.1115/1.4056534.

[15] Hwang, M.-h., Cha, H.-R., and Jung, S. Y., "Practical Endurance Estimation for Minimizing Energy Consumption of Multirotor Unmanned Aerial Vehicles," *Energies*, Vol. 11, No. 9, 2018. https://doi.org/10.3390/en11092221, URL https://www.mdpi.com/1996-1073/11/9/2221.

[16] Tamke, F., and Buscher, U., "The vehicle routing problem with drones and drone speed selection," *Computers & Operations Research*, Vol. 152, 2023, p. 106112. https://doi.org/https://doi.org/10.1016/j.cor.2022.106112, URL https://www.sciencedirect.com/science/article/pii/S0305054822003422.

[17] D'Andrea, R., "Guest Editorial Can Drones Deliver?" *IEEE Transactions on Automation Science and Engineering*, Vol. 11, No. 3, 2014, pp. 647–648. https://doi.org/10.1109/TASE.2014.2326952.

[18] Dorling, K., Heinrichs, J., Messier, G. G., and Magierowski, S., "Vehicle Routing Problems for Drone Delivery," *IEEE Transactions on Systems, Man, and Cybernetics: Systems*, Vol. 47, No. 1, 2017, p. 70 – 85. https://doi.org/10.1109/TSMC.2016.2582745, cited by: 761; All Open Access, Green Open Access.

[19] Lin, X., Perez, H. E., Mohan, S., Siegel, J. B., Stefanopoulou, A. G., Ding, Y., and Castanier, M. P., "A lumped-parameter electro-thermal model for cylindrical batteries," *Journal of Power Sources*, Vol. 257, 2014, pp. 1–11.

[20] Johnson, W., *Helicopter Theory*, Dover Publications, New York, NY, 1994.

[21] Michel, N., Sinha, A. K., Kong, Z., and Lin, X., "Multiphysical Modeling of Energy Dynamics for Multirotor Unmanned Aerial Vehicles," *2019 International Conference on Unmanned Aircraft Systems (ICUAS)*, IEEE, 2019, pp. 738–747.